\documentclass[a4paper,11pt]{article}

\usepackage{times}
\usepackage{natbib}
\usepackage[margin=25mm]{geometry}

\usepackage{helvet}
\usepackage{courier}
\usepackage[T1]{fontenc}
\usepackage{upquote}
\usepackage{examples}
\usepackage{cgloss4e}
\usepackage{balance}

\usepackage{verbatim}
\usepackage{cprotect}
\usepackage{url}
\usepackage{listings}

\usepackage{times}
\usepackage{url}
\usepackage{latexsym}
\usepackage{amsmath}
\usepackage[pdftex]{graphicx}
\usepackage{dblfloatfix}
\usepackage{graphics}
\usepackage[usenames,dvipsnames,svgnames,table]{xcolor}

\usepackage{amssymb,amsmath,epsfig}
\usepackage{mathpartir}
\usepackage{proof}
\usepackage{amsthm}
\usepackage{xspace}
\usepackage{algorithm}
\usepackage[noend]{algpseudocode}
\algrenewcommand{\algorithmicindent}{1em}

\newcommand{\notes}[1]{}

 \theoremstyle{definition}
 
\theoremstyle{plain}

\newcommand{\ith}[1]{\ensuremath{i^{{th}}}}

\renewcommand{\to}{{\rightarrow}}

\newcount\permx
\newcount\permy
\def\permdot#1#2{
\permx=#1 \advance\permx by-1
\permy=#2 \advance\permy by-1
\psframe[fillcolor=black, fillstyle=solid]
(\permx,\permy)(#1, #2)
}

\newcommand{\argmax}{\operatornamewithlimits{\mathbf{argmax}}}

\newcommand{\boxnum}[1]{{\setlength{\fboxsep}{1pt}\raisebox{1pt}{\hspace{1pt}\fbox{\tiny #1}\hspace{1pt}}}}
\newcommand{\ind}[1]{\ensuremath{^{\kern-0.5pt\boxnum{#1}}}}

\newcommand{\vecw}{\ensuremath{\mathbf{w}}\xspace}

\renewcommand{\root}{\ensuremath{\mathit{root}}\xspace}

\providecommand{\namecite}[1]{\citeauthor{{#1}} (\citeyear{{#1}})}

\newcommand{\smallnt}[1]{\ensuremath{_{\mbox{\tiny PP}}}\xspace}

\newcommand{\shift}{\ensuremath{\mathsf{sh}}\xspace}
\newcommand{\reduce}{\ensuremath{\mathsf{re}}\xspace}

\newcommand{\rreduce}{\ensuremath{\mathsf{re_{\small \curvearrowright}}}\xspace}

\newcommand{\pseudocode}{Algorithm}
\floatname{algorithm}{\pseudocode}

\iffalse

\else

\fi

\newcommand{\bPhi}{\ensuremath{\mathbf{\Phi}}\xspace}
\newcommand{\dPhi}{\ensuremath{\Delta\bPhi}\xspace}

\providecommand{\card}[1]{\lvert#1\rvert}  

\newcommand{\defeq}{\ensuremath{\stackrel{\Delta}{=}}\xspace}

\newcommand{\good}{\ensuremath{\mathit{good}}\xspace}
\newcommand{\bad}{\ensuremath{\mathit{bad}}\xspace}

\newcommand{\subtype}{\ensuremath{<:}}
\newcommand{\type}[1]{\ensuremath{\mathsf{{#1}}\xspace}}

\newcommand{\geoquery}{\ensuremath{\textsc{GeoQuery}}\xspace}
\newcommand{\jobs}{\ensuremath{\textsc{Jobs}}\xspace}
\newcommand{\atis}{\ensuremath{\textsc{Atis}}\xspace}
\newcommand{\tv}[1]{\ensuremath{\mbox{\tt \textquotesingle}\!\text{\tt{#1}}}}
\newcommand{\tva}{\ensuremath{\mbox{\tt \textquotesingle}\!\text{\tt{a}}}}

\newcommand{\pred}[1]{\ensuremath{\text{\bf{#1}}}}

\newcommand{\typetop}{\ensuremath{\type{top}}}
\newcommand{\typest}{\ensuremath{\type{st}}}
\newcommand{\typect}{\ensuremath{\type{ct}}}
\newcommand{\typelo}{\ensuremath{\type{lo}}}
\newcommand{\typerv}{\ensuremath{\type{rv}}}
\newcommand{\typelk}{\ensuremath{\type{lk}}}
\newcommand{\typeau}{\ensuremath{\type{au}}}
\newcommand{\typenu}{\ensuremath{\type{nu}}}
\newcommand{\typee}{\ensuremath{\type{e}}}
\newcommand{\typei}{\ensuremath{\type{i}}}
\newcommand{\typet}{\ensuremath{\type{t}}}
\newcommand{\typeT}{\ensuremath{{\sf T}}}
\newcommand{\typeS}{\ensuremath{{\sf S}}}
\newcommand{\typecol}{\ensuremath{:}\xspace}

\newcommand{\TISP}{\ensuremath{\text{\sc Tisp}}\xspace}

\usepackage{multirow}
\usepackage{tikz}
\usepackage{tikz-qtree}
\usetikzlibrary{arrows}

\title{Type-Driven Incremental Semantic Parsing with Polymorphism
}

\author{Kai Zhao\\
Graduate Center\\
City University of New York\\
{\tt kzhao.hf@gmail.com}
\and
Liang Huang\\
Queens College\\
City University of New York\\
{\tt liang.huang.sh@gmail.com}
}

\date{}

\begin{document}
\maketitle
\begin{abstract}

Semantic parsing has made significant progress, but most current semantic parsers 
are extremely slow (CKY-based) and rather primitive in representation. 
We introduce three new techniques to tackle these problems. 
First, we design the first linear-time incremental shift-reduce-style
semantic parsing algorithm
which is more efficient than conventional cubic-time bottom-up semantic parsers.
Second, our parser, being type-driven instead of syntax-driven,
uses type-checking to decide the direction of reduction,
which eliminates the need for a syntactic grammar such as CCG.
Third, 
to fully exploit the power of type-driven semantic parsing
beyond simple types (such as entities and truth values),
we borrow from programming language theory
the concepts of subtype polymorphism and parametric polymorphism
to enrich the type system in order to better guide the parsing.  
Our system learns very accurate parses in \geoquery, \jobs and \atis domains.
\end{abstract}

\section{Introduction}
\label{sec:intro}

Most existing semantic parsing efforts employ 
a CKY-style bottom-up parsing strategy 
to generate a meaning representation 
in simply typed lambda calculus \cite{zettlemoyer+collins:2005,lu+ng:2011}
or its variants \cite{wong+mooney:2007,liang+:2011}.
Although these works led to fairly accurate semantic parsers, 
there are two major drawbacks:
efficiency and expressiveness.


First, as many researches in syntactic parsing \cite{nivre:2008,zhang+clark:2011} have shown, 
compared to cubic-time CKY-style parsing,
incremental parsing can achieve comparable accuracies 
while being linear-time, which means orders of magnitude faster in practice.
We therefore introduce the first incremental parsing algorithm for semantic parsing.
More interestingly, unlike syntactic parsing,
our incremental semantic parsing algorithm, being strictly {\bf type-driven}, 
directly employs type checking
to automatically determine the direction of function application on-the-fly,
thus reducing the search space and eliminating the need for a syntactic grammar such as CCG
which explicitly encodes the direction of function application.


However, to fully exploit the power of type-driven incremental parsing,
we need a more sophisticated type system than simply typed lambda calculus.
We argue that it is beneficial to incorporate an explicit subtype hierarchy,
such that ambiguous terms 
can be grounded based on context in a more explicit and declarative fashion. 
Compare the following two phrases:
\begin{examples}
\item the mayor of New York?
\item the capital of New York?
\end{examples}
If we know that {\em mayor} is a function from city to person,
then the first New York can only be of type city;
similarly knowing {\em capital} maps states to cities disambiguates
the second New York to be of type state.
This can not be done using a simple type system with just entities and booleans.

Now let us consider a more complex question which will be our running example
in this paper:
\begin{examples}
\item What is the capital of the largest state by area?
\end{examples}
Since we know {\em capital} takes a state as input,
we expect {\em the largest state by area} to return a state.
But does {\em largest} always return a state type? 
Notice that it is polymorphic,
for example, {\em largest city by population}, or
{\em largest lake by perimeter}.
So there is no unique type for {\em largest}: its return type should depend on the type of 
its first argument ({\em city}, {\em state}, or {\em lake}).
This observation motivates us to introduce the powerful mechanism of parametric polymorphism
from programming languages into the type system for natural language.
For example, we can define the type of {\em largest} to be a template
\[ \pred{largest}:(\tv{a}\to\typet)\to(\tv{a}\to\typei)\to\tv{a} \]
where \tv{a} is a {\em type variable} that can match any type
(for formal details see Section~\ref{sec:lambda}).

Just like in functional programming languages such as ML or Haskell, 
type variables can be bound to a real type
(or a range of types) during function application, using the technique of type inference.
In the above example, when {\em largest} is applied to {\em city},
we know that type variable \tv{a} is bound to type city (or its subtype),
so that {\em largest} would eventually return a city.

We make the following contributions:
\begin{itemize}
  \item We design a linear-time incremental semantic parsing algorithm (Section~\ref{sec:parsing}),
    which is much more efficient than the majority of existing semantic parsers
    that are cubic-time CKY-based.
  \item In line with classical Montague theory \cite{heim+kratzer:1998},
  our parser is type-driven parsing instead of syntax-driven as in CCG-based efforts
  \cite{zettlemoyer+collins:2005,kwiatkowski+:2011,krishnamurthy+mitchell:2014} (Section~\ref{sec:reduce}). 
  \item We introduce parametric polymorphism into natural language semantics (Section~\ref{sec:lambda}), along with proper treatment of subtype polymorphism, and 
    implement Hindley-Milner style type inference \cite[Chap.~10]{pierce:2005}
     during parsing (Section~\ref{sec:parse2}).\footnote{There
are three kinds of polymorphisms in programming languages:
parametric (e.g., C++ templates), subtyping, and ad-hoc (e.g., operator overloading).
See \cite[Chap.~15]{TAPL} for details.}

  \item We adapt the latent-variable max-violation perceptron training from machine translation \cite{yu+:2013},
  which is a perfect fit for semantic parsing due to its huge search space (Section~\ref{sec:train}).
\end{itemize}

Experiments on \geoquery, \jobs and \atis domains
show close to state-of-the-art performances,
and demonstrate the advantage of a powerful type system.

\section{Type-Driven Incremental Parsing}
\label{sec:parsing}


We start with the simplest meaning representation (MR), {\em untyped lambda calculus},
and then introduce typing and the incremental parsing algorithm for it.
Later in Section~\ref{sec:lambda}, 
we add subtyping and type polymorphism 
to enrich the system. 

\subsection{Meaning Representation with Types}

The untyped MR for the running example is:
\vspace{0.2cm}
\begin{center}
\hspace{-0.5cm}
\begin{tabular}{l}
Q:  What is the capital of the largest state by area? \\[0.2cm]
MR:  
$(\pred{capital}\quad (\pred{argmax}\quad  \pred{state}\quad 
 \pred{size}))$
\end{tabular}
\end{center}

Note the binary function \pred{argmax}$(\cdot,\cdot)$
is a higher-order function that 
takes two other functions as input: 
the first argument is a ``domain'' function that defines the set to search for,
and second argument is an ``evaluation'' function that returns a integer for an 
element in that domain.
In other words 
\[\displaystyle \pred{argmax}(f, g) = \argmax_{x:f(x)} g(x).\]

The simply typed lambda calculus \cite{heim+kratzer:1998,lu+ng:2011}
augments the system with types, including base types (entities \typee, truth values \typet,
or numbers \typei), and function types (e.g., $\typee\to\typet$).
So function \pred{capital} is of type $\typee\to\typee$, \pred{state} is of type $\typee\to\typet$,
and \pred{size} is of type $\typee\to\typei$.
The \pred{argmax} function is of type $(\typee\to\typet)\to(\typee\to\typei)\to\typee$.\footnote{Note that
the type notation is always {\em curried}, i.e., we
represent a binary function as a unary function that returns another unary function.
Also the type notation is always {\em right-associative},
so $(\typee\to\typet)\to((\typee\to\typei)\to\typee)$ is also written as
$(\typee\to\typet)\to(\typee\to\typei)\to\typee$.}
The simply typed MR is now written as
\begin{align*}
(\pred{capital}\!:\!\typee\to\typee \quad
(\pred{argmax}\!: & (\typee\to\typet)\to(\typee\to\typei)\to\typee \\
             & \pred{state}\!:\!\typee\to\typet\;\; \pred{size}\!:\!\typee\to\typei))
).
\end{align*}

\begin{figure*}[t]
\centering
\resizebox{0.7\textwidth}{!}{
\begin{tabular}{cl|l|l}
step & action & stack after action & queue\\
\hline
0 & - & $\phi$   & what ... \\
\hline
1--3 & skip & $\phi$& capital ... \\
\hline
4 & $\shift_\text{capital}$ & \pred{capital}:$\typee\to\typee$ & of ...\\
\hline
7 & $\shift_\text{largest}$ & \pred{capital}:$\typee\to\typee$ \qquad \pred{argmax}:$(\typee\to\typet)\to(\typee\to\typei)\to\typee$ & state ...\\
\hline
8 & $\shift_\text{state}$   & \pred{capital}:$\typee\to\typee$ \qquad \pred{argmax}:$(\typee\to\typet)\to(\typee\to\typei)\to\typee$ \qquad \pred{state}:$\typee\to\typet$ & by ...\\
\hline
9 & $\rreduce$              & \pred{capital}:$\typee\to\typee$ \qquad (\pred{argmax} \pred{state}):$(\typee\to\typei)\to\typee$ & by ...\\
\hline
11 & \shift$_\text{area}$   & \pred{capital}:$\typee\to\typee$ \qquad (\pred{argmax} \pred{state}):$(\typee\to\typei)\to\typee$ \qquad\quad\ \pred{size}:$\typee\to\typei$ & ?\\
\hline
12 & $\rreduce$              & \pred{capital}:$\typee\to\typee$ \qquad (\pred{argmax} \pred{state} \pred{size}):$\typee$ & ? \\
\hline
13 & $\rreduce$              & (\pred{capital} (\pred{argmax} \pred{state} \pred{size})):$\typee$ & ? \\
\end{tabular}
}
\\[0.3cm]
(a) type-driven incremental parsing with simple types  (entities \typee, truth values \typet, and integers \typei); see Section~\ref{sec:parsing}. \\[0.7cm]

\resizebox{0.99\textwidth}{!}{
\begin{tabular}{cl|l|l|l}
step & action & stack after action & queue & typing\\
\hline
0 & - & $\phi$   & what... \\
\hline
1--3 & skip & $\phi$& capital... \\
\hline
4 & $\shift_\text{capital}$ & \pred{capital}:$\typest\to\typect$ & of...\\
\hline
7 & $\shift_\text{largest}$ & \pred{capital}:$\typest\to\typect \quad \pred{argmax}\!:\!(\tva\to\typet)\to(\tva\to\typei)\to\tva$ & state...\\
\hline
8 & $\shift_\text{state}$   & \pred{capital}:$\typest\to\typect \quad \pred{argmax}\!:\!(\tva\to\typet)\to(\tva\to\typei)\to\tva \quad \pred{state}\!:\!\typest\to\typet$ & by...\\
\hline
9 & $\rreduce$              & \pred{capital}:$\typest\to\typect \quad (\pred{argmax}\ \ \pred{state})\!:\!(\typest\to\typei)\to\typest$ & by... & binding: $\tva=\typest$\\
\hline
11 & \shift$_\text{area}$   & \pred{capital}:$\typest\to\typect \quad (\pred{argmax}\ \ \pred{state})\!:\!(\typest\to\typei)\to\typest \ \quad\quad\, \pred{size}\!:\!\typelo\to\typei$ & ?\\
\hline
12 & $\rreduce$              & \pred{capital}:$\typest\to\typect \quad (\pred{argmax}\ \ \pred{state}\ \ \pred{size})\!:\!\typest$ & ? & $\typest\subtype\typelo\Rightarrow (\typelo\to\typei)\!\subtype\!(\typest\to\typei)$\\
\hline
13 & $\rreduce$              & $(\pred{capital}\ \ (\pred{argmax}\ \ \pred{state}\ \ \pred{size}))\!:\!\typect$ & ? \\
\end{tabular}
}
\\[0.3cm]
(b) type-driven incremental parsing with subtyping (\subtype) and type polymorphism (e.g., type variable \tv{a}); see Section~\ref{sec:parse2}.

\caption{Type-driven Incremental Semantic Parsing (\TISP) with (a) simple types and (b) subtyping+polymorphism  on
the example question:
``what is the capital of the largest state by area?''.
Steps 5--6 and 10 are skip actions and thus omitted.
The stack and queue in each row are the results {\it after} each action.\label{fig:parse}}
\end{figure*}


\subsection{Incremental Semantic Parsing: An Example}

We use the above running example to explain 
our type-driven incremental semantic parsing algorithm.
Figure~\ref{fig:parse} (a) illustrates the full derivation.

Similar to a standard shift-reduce parser,
we maintain a {\it stack} and a {\it queue}.
The queue contains words to be parsed,
while the stack contains subexpressions of the final MR,
where each subexpression is a valid typed lambda expression.
At each step, the parser choose to {\bf shift} or {\bf reduce},
but unlike standard shift-reduce parser,
there is also a third possible action, {\bf skip},
which skips a semantically vacuous word (e.g., ``the'', ``of'', ``is'', etc.).
For example, 
the first three words of the example question {\em ``What is the ...''}
 are all skipped
({\bf steps 1--3} in Figure~\ref{fig:parse} (a)).




The parser then {\bf shifts} the next word, {\em ``capital''}, 
from the queue to the stack.
But unlike incremental syntactic parsing where the word itself is moved onto the stack,
here we need to find a {\bf grounded} predicate in the GeoQuery domain
for the current word.
In this example we find the predicate:
\[\pred{capital}\!:\!\typee\to\typee\]
and put it on the stack ({\bf step 4}).

Next, words {\em ``of the''} are skipped (steps 5--6).
Then for word {\em ``largest''}, we shift 
the predicate
\[\pred{argmax}\!:\!(\typee\to\typet)\to(\typee\to\typei)\to\typee \]
onto the stack ({\bf step 7}), which becomes
\[\pred{capital}\!:\!\typee\to\typee \quad \pred{argmax}\!:\!(\typee\to\typet)\to(\typee\to\typei)\to\typee. \]

At this step we have two expressions on the stack and we could attempt to reduce.
But type checking fails because 
for left reduce, \pred{argmax} expects an argument (its ``domain'' function) 
of type $(\typee\to\typet)$
which is different from \pred{capital}'s type $(\typee\to\typee)$,
so is the case for right reduce.

So we have to shift again. This time for word {\em ``state''} we shift the predicate
\[ \pred{state}\!:\!\typee\to\typet \]
onto the stack, which becomes:
\[\pred{capital}\!:\!\typee\to\typee \quad \pred{argmax}\!:\!(\typee\to\typet)\to(\typee\to\typei)\to\typee\quad  \pred{state}\!:\!\typee\to\typet. \]
\subsection{Type-Driven Reduce}
\label{sec:reduce}

At this step we can finally perform a {\bf reduce} action,
since the top two expressions on the stack 
pass the type-checking for rightward function application
(a partial application):
\pred{argmax} expects an $(\typee\to\typet)$ argument,
which is exactly the type of \pred{state}.
So we conduct a right-reduce, 
applying \pred{argmax} on \pred{state},
and the resulting expression is:
\[
(\pred{argmax}\quad \pred{state})\!:\!(\typee\to\typei)\to\typee
\]
while the stack becomes ({\bf step 9})
\[
\pred{capital}\!:\!\typee\to\typee \qquad (\pred{argmax}\quad \pred{state})\!:\!(\typee\to\typei)\to\typee
\]
Now if we want to continue reduction,
it does not type check for either left or right reduction,
so we have to shift again.

So we move on to shift the final word {\em ``area''}
with the grounded predicate in GeoQuery database:
\[
\pred{size}\!:\!\typee\to\typei
\]
and the stack becomes ({\bf step 11}):
\[
\pred{capital}\!:\!\typee\to\typee \qquad (\pred{argmax}\quad \pred{state})\!:\!(\typee\to\typei)\to\typee
\quad
\pred{size}\!:\!\typee\to\typei.
\]
Now apparently we can do a right reduce supported by type checking ({\bf step 12}):
\[
\pred{capital}\!:\!\typee\to\typee \qquad (\pred{argmax}\quad \pred{state} \quad \pred{size}) \!:\!\typee
\]
followed by another, final, right reduce ({\bf step 13}):
\[
(\pred{capital} \qquad (\pred{argmax}\quad \pred{state} \quad \pred{size})) \!:\!\typee.
\]



Here we can see the novelty of our shift-reduce parser:
its decisions are largely driven by the type system.
When we attempt a reduce, {\bf at most} one of the two reduce actions (left, right)
is possible thanks to type checking,
and when neither is allowed, we have to shift (or skip).
This observation suggests that our incremental parser is more 
deterministic than those syntactic incremental parsers
whose each step always faces a three-way decision (shift, left-reduce, right-reduce). 
We also note that this type-checking mechanism,
inspired by 
the classical type-driven theory in linguistics
\cite{heim+kratzer:1998},
eliminates the need for an {\it explicit} encoding of direction
as in CCG, which makes our formalism much simpler than the 
synchronous syntactic-semantic ones in most other 
semantic parsing efforts \cite{zettlemoyer+collins:2005,zettlemoyer+collins:2007,wong+mooney:2007}.



\bigskip

As a side note, besides function application, reduce also occurs when the top two expressions on the stack can be combined to represent a more specific meaning,
which we call {\it union}. 

For example, when parsing the phrase ``major city'', 
we have the top two expressions on the stack
\[\pred{major}\!:\!\typee\to\typet \quad \pred{city}\!:\!\typee\to\typet\]
We can combine the two expressions using predicate \pred{and} 
since their types match, and get 
\[\lambda x\!:\!\typee\ .\ (\pred{and}\!:\!\typet\to\typet\to\typet\ (\pred{major}\!:\!\typee\to\typet\ \ x)\ 
                                                (\pred{city}\!:\!\typee\to\typet\ \ x)),\]
where type $\typet\to\typet\to\typet$ takes two booleans and return one
(again, using currying notation).

\section{Subtype and Type Polymorphisms}
\label{sec:lambda}




As mentioned in Section 1, simply typed lambda calculus representation 
can not distinguish between Mississippi the river
and Mississippi the state since they both have the same type \typee.
Furthermore, currently function \pred{capital} can apply to any entity type,
for example \pred{capital}(\pred{boston}),
which should have been disallowed by the type checker.
So we need a more sophisticated type system that helps ground terms to real-world entities,
and this refined type system will in turn help type-driven parsing.


\begin{figure}
\centering
\scalebox{.85}{
\begin{tikzpicture}
\tikzset{level distance=35pt, every tree node/.style={align=center,anchor=base}}
\begin{scope}
\Tree 
[.{\typetop\\ root type}
  [.{\typelo\\ location}
    [.{\typeau\\ admin. unit}
      {\typest\\ state}
      {\typect\\ city}
    ]
    [.{\typenu\\ nature unit}  
      {\typerv\\ river}
      {\typelk\\ lake}
    ]  
  ]
  {\typei\\ integer}
]
\end{scope}

\begin{scope}[shift={(3,0)}]
\Tree 
[.{\typet\\boolean}
]
\end{scope}
\end{tikzpicture}
}
\caption{Type hierarchy for \geoquery domain (slightly simplified for presentation).
\label{fig:typehier}}
\end{figure}
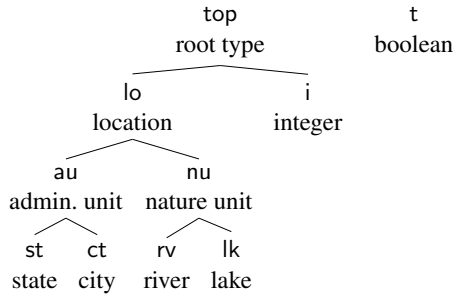

\subsection{Augmenting MR with Subtyping}

We first augment the meaning representation with a type hierarchy 
which is domain specific.
For example Figure~\ref{fig:typehier} shows 
a (slightly simplified) version of the type hierarchy for \geoquery domain.
Here the root type $\typetop$
has a subtype of locations, \typelo,
which consists of two different kinds of locations,
administrative units (\typeau) including
states (\typest) and cities (\typect),
and nature units (\typenu) including 
rivers (\typerv) and lakes (\typelk).
We use \subtype\/ to denote the (transitive, reflexive, and antisymmetric) {\bf subtyping relation}
between types;
for example in \geoquery we have
$\typest \subtype \typelo$, $\typerv \subtype \typenu$, 
and $\typeT \subtype \typetop$ for any type $\typeT$.

In addition we have an integer type $\typei$ derived from the root type \typetop.
The boolean type $\typet$ does not belong to the type hierarchy,
because it does not represent the semantics from the task domain.

Each constant in the \geoquery domain is well typed.
For example, there are states 
(
\pred{mississippi}:\typest),
cities 
(
\pred{boston}:\typect),
rivers (\pred{mississippi}:\typerv
),
and lakes (\pred{tahoe}:\typelk
).
Note that the names like {\it mississippi} appears twice for two different entities.
The fact that we can distinguish them by type is a crucial advantage of
a typed semantic formalism.

Similarly each predicate is also typed.
For example, we can query 
the length of a river, 
\pred{len}:{\typerv$\to$\typei},
or the population of some administrative unit,
\pred{population}:{\typeau$\to$\typei}.
Notice that \pred{population}$(\cdot)$ can be applied 
to both states and cities, since they are subtypes of administrative unit, i.e., $\typest \subtype \typeau$
and $\typect \subtype \typeau$.
This is because, as in Java and C++,
a function 
that expects a type $\typeT$ argument can always take an argument of another type $\typeS$
which is a subtype of $\typeT$. More formally:
\begin{equation}
\inferrule{ e_2:\typeS \qquad \typeS \subtype \typeT}
{(\lambda x:\!\typeT\ .\ e_{1})\;\; e_2\ \to\ [x \mapsto e_2] e_{1}} ,
\label{eq:app}
\end{equation}
where $[x\mapsto\!e_2] e_1$ means substituting all occurrences of variable $x$ in expression $e_1$ with expression $e_2$.
For example, we can query whether two locations are adjacent, using
\pred{next\_to}:{\typelo$\to$(\typelo$\to$\typet)},
and similarly the \pred{next\_to}$(\cdot,\cdot)$ function
can be applied to two states, or to a river and a city, etc.


The above type system works smoothly for first-order functions (i.e., predicates taking atomic type arguments),
but the situation with higher-order functions (i.e., predicates that take functions as input) is more involved.
What is the type of \pred{argmax}? One possibility is to define it to be as general as possible,
as in the simply typed version (and many conventional semantic parsers):
\[\pred{argmax}\!:\!(\typetop\to\typet)\to(\typetop\to\typei)\to\typetop.\]
But this actually no longer works for our sophisticated type system
for the following reason.

Intuitively, remember 
that \pred{capital}:$\typest\to\typect$ is now a function that takes a state as input,
so the return type of \pred{argmax} must be a state or its subtype,
rather than $\typetop$ which is a supertype of $\typest$.
But we can not simply replace $\typetop$ by $\typest$, since \pred{argmax} can also be applied in other
scenarios such as ``the largest city'' or ``the longest river''.
In other words, \pred{argmax} is a {\em polymorphic} function,
and to assign a correct type for it we have to introduce {\it type variables}
(widely used in functional programming languages such as Haskell and ML,
and also in C++ templates).
We define 
\[ \pred{argmax}\!:\!(\tv{a}\to\typet)\to(\tv{a}\to\typei)\to\tv{a} \]
where the type variable \tv{a} is a place-holder for ``any type''.

Before we move on, there is an important consequence of polymorphism
worth mentioning here.
For the types of unary predicates
such as 
\pred{city}($\cdot$) and \pred{state}($\cdot$) that 
{\it characterize} its argument,
we define theirs argument types to be the required type, i.e.,
$\pred{city}\!:\!\typect\to\typet$, and $\pred{state}\!:\!\typest\to\typet$.
This might look a little weird
since everything in the domain of those functions are always mapped to true;
i.e., $f(x)$ is either undefined or true, and never false for such $f$'s.
This is different from classical simply-typed Montague semantics \cite{heim+kratzer:1998}
which defines such predicates as type $\typetop\to\typet$ so that 
$\pred{city}(\pred{mississippi}\!:\!\typest)$ returns false.
The reason for our design is, again, due to subtyping and polymorphism:
\pred{capital} takes a state type as input,
so \pred{argmax} must returns a state,
and therefore its first argument, the \pred{state} function, must have type $\typest\to\typet$
so that the matched type variable \tv{a} will be bound to \typest.
This more refined design will also help prune unnecessary argument matching
using type checking.


\subsection{Parsing with Subtype Polymorphism and Parametric Polymorphism}
\label{sec:parse2}

We modify the previous incremental parsing algorithm with simple types  (Section~\ref{sec:parsing}) %
to accommodate subtyping and polymorphic types.
Figure~\ref{fig:parse} (b) shows 
the derivation of the running example using the new parsing algorithm. 
Below we focus on the differences brought by the new algorithm.



In step 4, unlike $\pred{capital}:\typee\to\typee$, we shift the predicate
\[
\pred{capital}:\typest\to\typect
\]
and in step 7,
we shift the polymorphic expression for {\em ``largest''}
\[
\pred{argmax}\!:\!(\tva\to\typet)\to(\tva\to\typei)\to\tva \label{eq:argmax}
\]
And after the shift in step 8, the stack becomes
\[
\pred{capital}\!:\!\typest\to\typect \quad \pred{argmax}\!:\!(\tva\to\typet)\to(\tva\to\typei)\to\tva \quad \pred{state}\!:\!\typest\to\typet
\]

At {\bf step 9}, in order to apply \pred{argmax} onto
\(\pred{state}\!:\!\typest\to\typet\),
we simply {\bf bind} type variable $\tva$ to type $\typest$,
i.e.,
\[
\pred{argmax}\!: (\typest\to\typet)\to(\typest\to\typei)\to\typest \quad
\pred{state}\!:\!\typest\to\typet 
\]
results in
\[
(\pred{argmax} \;\;
\pred{state}):(\typest\to\typei)\to\typest 
\]

After the shift in {\bf step 11},
the stack becomes:
\[
\pred{capital}\!:\!\typest\to\typect \quad
(\pred{argmax} \;\; \pred{state})\!:\!(\typest\to\typei)\to\typest 
\quad \pred{size}\!:\!\typelo\to\typei.
\]
Can we still apply right reduce here?
According to the subtyping rule (Eq.~\ref{eq:app}),
we want 
\[
\typelo\to\typei\subtype\typest\to\typei
\]
to hold, knowing that $\typest\subtype\typelo$.
Luckily, there is a rule about function types in type theory that exactly fits here:
\begin{equation}
\inferrule{{\sf A}\subtype {\sf B} } 
{{\sf B}\to{\sf C} \subtype {\sf A}\to{\sf C}}
\label{eq:contra}
\end{equation}
which states the input side is reversed (contravariant).
This might look counterintuitive at the first glance, but the intuition is that, 
it is safe to allow the function $\pred{size}$ of type $\typelo\to\typei$ to be used 
in the context where another type $\typest\to\typei$ is expected,
since in that context the argument passed to $\pred{size}$ will be state type (\typest),
which is a subtype of location type ($\typelo$) that $\pred{size}$ expects,
which in turn will not surprise $\pred{size}$. 
See the classical type theory textbook \cite[Chap.~15.2]{TAPL} for details.
See Figure~\ref{fig:parse} (b) for the full derivation.

\section{Training: Latent Variable Perceptron}
\label{sec:train}

We follow the Latent Variable Violation-Fixing Perceptron framework \cite{huang+:2012,yu+:2013} for the training. 

\subsection{Framework}


The key challenge in the training is that, 
for each question, there might be many different unknown derivations that lead to its annotated MR,
which is known as the {\it spurious ambiguity}.
In our type-driven incremental semantic parsing task,
the spurious ambiguity is caused by 
how the expression templates are chosen and grounded during the shift step,
and the different reduce orders that lead to the same result.
We treat this unknown information as latent variable.

More formally, we denote   $D(x)$ to be the set of {\it all}  
partial and full parsing derivations for an input sentence $x$,
and  $\mathit{mr}(d)$ to be  
the MR yielded by a full derivation $d$. 
Then we define the sets of (partial and full) reference derivations as:
\begin{align*}
\good_i(x,y) \defeq \{d\in D(x) \mid \  &\card{d}=i, \exists \text{full derivation } d' \text{ s.t. } \\
&d \text{ is a prefix of } d', \mathit{mr}(d')=y\},
\end{align*}

Those ``bad'' partial and full derivations that do not lead to the annotated MR can be defined as:
\[\bad_i(x,y) \defeq \{d\in D(x) \mid d\not\in \good_i(x,y), \card{d}=i\}.\]

At step $i$, the best reference partial derivation is
\begin{equation}
d_i^+(x, y) \defeq \argmax_{d\in \good_i(x,y)} \vecw\cdot\bPhi(x,d), \label{eq:goldderiv}
\end{equation}
while the Viterbi partial derivation is
\begin{equation}
d_i^-(x, y) \defeq \argmax_{d\in\bad_i(x,y)} \vecw\cdot\bPhi(x,d), \label{eq:viterbideriv}
\end{equation}
where $\bPhi(x, d)$ is the defined feature set for derivation $d$.

In practice, to compute Eq.~\ref{eq:viterbideriv} exactly is intractable, 
and we resort to beam search. 

Following \namecite{yu+:2013}, we then find the step $i^*$ with the maximal score difference between 
the best reference partial derivation and the Viterbi partial derivation:
\[i^* \defeq \argmax_i\vecw\cdot\dPhi(x, d_i^+(x,y), d_i^-(x,y)),\]  
and do update:
\[\vecw \gets \vecw + \dPhi(x, d_{i^*}^+(x,y), d_{i^*}^-(x,y))\]
where $\dPhi(x, d, d') \defeq \bPhi(x, d) - \bPhi(x, d')$.

\subsection{Forced Decoding}
We use forced decoding to retrieve the reference derivations $\good_i(x,y)$
for each question/MR pair $(x,y)$ in Eq.~\ref{eq:goldderiv}.

Unlike syntactic incremental parsing, 
where the forced decoding can be done in polynomial time \cite{goldberg+:2014},
we do not have an algorithm designed for efficient forced decoding.
We apply exponential-time brute-force search to calculate $\good(x,y)$,
during which we do pruning based on the predicate application orders.

However, this requires heavy computation we can not afford.
In practice we choose {\it multi-pass} forced decoding.
First we use brute-force search to decode, but with a time limit. 
Then we train a Perceptron using successfully decoded reference derivations,
and use the trained Perceptron to decode the unfinished questions
with a large beam. We then add the reference derivations newly discovered
into the next step training. 

\section{Experiments}
\label{sec:exps}

We implement our type-driven incremental semantic parser (\TISP) using {\tt Python},
and evaluate its performance of both speed and accuracy on \geoquery and \jobs
datasets.


Our feature design is inspired by the very effective Word-Edge features 
in syntactic parsing \cite{charniak+johnson:2005}
and MT \cite{cars2:2008}.
From each parsing state, we collect atomic features
including the types 
and the leftmost and rightmost words 
of the span of the top 3 MR expressions on the stack,
the top 3 words on the queue,
the grounded predicate names and the ID of the expression template used in the shift action.

To ease the overfitting problem caused by the feature sparsity,
we assign different budgets to different kinds of features
and only generate feature combinations within a budget limit.
We get 84 combined feature templates in total.

For evaluation, we follow \namecite{zettlemoyer+collins:2005} to use {\it precision} and {\it recall}, where
\[\text{Precision} = \frac{\# \text{ of correctly parsed questions}}{\# \text{ of successfully parsed questions}},\]
and 
\[\text{Recall} = \frac{\# \text{ of correctly parsed questions}}{\# \text{ of questions}}.\]

\subsection{Evaluation on \geoquery Dataset}
\label{sec:geoquery}

We first evaluate \TISP on \geoquery dataset.

Following the scheme of \namecite{zettlemoyer+collins:2007},
we use the first 600 sentences of Geo880 as the training set 
and the rest 280 sentences as the testing set.

Note that 
we do not have a separate development set,
due to the relatively small size of Geo880.
So to find the best 
number of iterations to stop the training,
we do a 10-fold cross-validation training over the training set,
and choose to train 20 iterations
and then evaluate.

We use two-pass forced decoding.
In the initial brute-force pass we set the time limit to 1,200 seconds,
and find the reference derivations for 530 of the total 600 training sentences,
a coverage of $\sim88\%$.
In the second pass we set beam size to 16,384 and get 581 sentences covered ($\sim97\%$).


In the training and evaluating time,
we use a very small beam size of 16,
which gives us very fast decoding.
In serial mode, 
our parser takes $\sim\!83$s to decode 
the 280 sentences (2,147 words) in the testing
set, which means $\sim\!0.3$s per sentence, or $\sim\!0.04$s per word.

We compare the our accuracy performance with existing methods in Table~\ref{tab:eval}.
Given that all other methods use CKY-style parsing,
our method is well balanced between accuracy and speed.

\begin{table}
\centering
\begin{tabular}{l|ccc|ccc|ccc}
& \multicolumn{3}{c|}{\geoquery} & \multicolumn{3}{c|}{\jobs} & \multicolumn{3}{c}{\atis}\\
System & P & R & F1 & P & R & F1 & P & R & F1\\ 
\hline
Z\&C'05 & {\bf 96.3} & 79.3 & 87.0 & {\bf 97.3} & 79.3 & 87.4 & - &- &- \\
Z\&C'07 & 91.6 & 86.1  &  88.8 & - &- &-& {\bf 85.8} &{\bf 84.6} &{\bf 85.2}\\
UBL & 94.1 & 85.0 & 89.3 & - &- &-& 72.1 &71.4 &71.7\\
FUBL & 88.6 & 88.6 & 88.6& - &- &-& 82.8 & 82.8 &82.8\\
\hline
\TISP(simple type) & 89.7 & 86.8 & 88.2 & 76.4 & 76.4 & 76.4 & - &- &-\\
\TISP & 92.9 & {\bf 88.9} & {\bf 90.9}& 85.0 &{\bf 85.0} &{\bf 85.0} & 84.7 & 84.2 &84.4\\
\hline
$\lambda$-WASP$^\star$ & 92.0 & 86.6 & 94.1 & - &- &- & - &- &-
\end{tabular}
\caption{Performances (precision, recall, and F1) of various parsing algorithms on \geoquery, \jobs, and \atis datasets.
$^\star$: $\lambda$-WASP for \geoquery is trained on 792 examples.
\label{tab:eval}}
\end{table}

In addition, to unveil the helpfulness of our type system,
we train a parser with only simple types. (Table~\ref{tab:eval})
In this setting,
the predicates only have primitive types of location \type{lo},
integer \type{i}, and boolean \type{t},
while the constants still keep their types.
It still has the type system,
but it is weaker than the polymorphic one.
Its accuracy is lower than the standard one,
mostly caused by that the type system can not help 
pruning the wrong applications like
\[ (\pred{population}\text{:}\type{au}\to\type{i}\ \ \pred{mississippi}\text{:}\type{rv}).\]

\subsection{Evaluations on \jobs and \atis Datasets}

\begin{figure}
\centering
\scalebox{.95}{
\begin{tikzpicture}
\tikzset{level distance=35pt, every tree node/.style={align=center,anchor=base}}
\begin{scope}
\Tree 
[.{\type{top}\\root type}
  [.{\type{qa}\\ qualification}
    {\type{ye}\\ year}
    {\type{ar}\\ area}
    {\type{pa}\\ platform}  
  ]
  {\type{jb}\\ job}
  {\type{i}\\ integer}
]
\end{scope}
\begin{scope}[shift={(3,0)}]
\Tree 
[.{\typet\\boolean}
]
\end{scope}
\end{tikzpicture}
}
\caption{Type hierarchy for \jobs domain (slightly simplified for presentation).
\label{fig:typehierjobs}}
\end{figure}
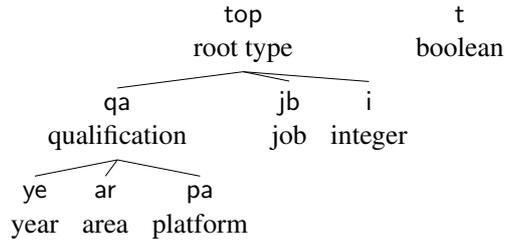

The \jobs domain contains descriptions about required and desired qualifications of a job.
The qualifications include programming language (\type{la}), 
years of experience (\type{ye}), 
diplomat degree (\type{de}), area of fields (\type{ar}), platform (\type{pa}),
title of the job (\type{ti}), etc.
We show a simplified version of the type hierarchy 
for \jobs in Figure~\ref{fig:typehierjobs}.

Following the splitting scheme of \namecite{zettlemoyer+collins:2005},
we use 500 sentences as training set and 140 sentences as testing set.

Table~\ref{tab:eval} shows that our algorithm achieves significantly higher recall than existing method of \namecite{zettlemoyer+collins:2005}, although our precision is not as high as theirs.
This is actually because our method parses a lot more questions in the dataset, 
as the column of the percentage of successfully parsed sentences suggests.

We also evaluate the performance of \TISP on \atis dataset as in Table~\ref{tab:eval}. 
\atis dataset contains more than 5,000 examples and is a lot larger than \geoquery and \jobs.
Our method achieves comparable performance on this dataset.
Due to space constraints, we do not show its type hierarchy here.

\section{Related Work}
\label{sec:related}


\namecite{zettlemoyer+collins:2005} introduce a type hierarchy to semantic parsing 
and parse with typed lambda calculus combined with CCG.
However, 
simply introducing subtyped predicates 
without polymorphism 
will cause type checking failures in handling high-order functions,
as shown in Section~\ref{sec:lambda}.
Furthermore, our system, being type-driven, almost completely rely on the types of MR expressions
to guide parsing (except for some simple POS tag triggers)
while their system is heavily CCG-based and syntax-driven.

\namecite{kwiatkowski+:2013} use ``on-the-fly'' matching to 
fetch the most possible predicate in the dataset for some MR subexpression.
The matching happens at the end of parsing, 
and is constrained by the type of the subexpression.
We do matching and parsing jointly, 
both of which are constrained by the typing, and affect the typing,
which is more similar to how human do semantic parsing,
i.e., we parse part of the sentence and bind that part to some specific meaning,
and continue parsing using grounded meaning.  

\namecite{wong+mooney:2007} also use type information to help reduce unnecessary 
tree joining in decoding.
However, their types are static,
while our type system is stronger so that we can infer type from polymorphism,
which gives use better search quality in decoding.

\section{Conclusions and Future Work}
We have presented an incremental semantic parser that is guided 
by a powerful type system of subtyping and parametric polymorphism. 
This polymorphism greatly reduced the number of templates
and effectively pruned search space during the parsing.
Our parser is competitive with state-of-the-art accuracies,
but, being linear-time, is orders of magnitude faster 
than CKY-based parsers in theory and in practice.

For future work,
we would like to work on weakly supervised learning 
that learn from question-answer pairs instead of question-MR pairs,
where the datasets are larger, 
and \TISP should benefit more on such problems.


\bibliographystyle{chicago}
\bibliography{thesis}

\end{document}